\documentclass[letterpaper]{article} % DO NOT CHANGE THIS
\usepackage{aaai2026}  % DO NOT CHANGE THIS
\usepackage{times}  % DO NOT CHANGE THIS
\usepackage{helvet}  % DO NOT CHANGE THIS
\usepackage{courier}  % DO NOT CHANGE THIS
\usepackage[hyphens]{url}  % DO NOT CHANGE THIS
\usepackage{graphicx} % DO NOT CHANGE THIS
\usepackage{natbib}  % DO NOT CHANGE THIS AND DO NOT ADD ANY OPTIONS TO IT
\usepackage{caption} % DO NOT CHANGE THIS AND DO NOT ADD ANY OPTIONS TO IT
\usepackage{algorithm}
\usepackage{algorithmic}
\usepackage{booktabs}
\usepackage{url}
\usepackage{graphicx}
\usepackage{float}

\usepackage{newfloat}
\usepackage{listings}
\DeclareCaptionStyle{ruled}{labelfont=normalfont,labelsep=colon,strut=off} % DO NOT CHANGE THIS
\floatstyle{ruled}
\newfloat{listing}{tb}{lst}{}
\floatname{listing}{Listing}
\usepackage{amsmath}
\usepackage{amssymb}

\title{Optimizing Against Safety Representations: Activation-Guided Adversarial Suffixes and the Geometry of Refusal}
\author{
    Ege Cakar, Hannah Guan, Kayden Kehe
}
\affiliations{
    \textsuperscript{}Harvard University\\
    \textsuperscript{}John A. Paulson School of Engineering and Applied Sciences\\
    \{ecakar, hguan, kaydenkehe\}@college.harvard.edu
}

\begin{document}
\frenchspacing
\nocopyright
\maketitle

\begin{abstract}
Behavioral alignment in large language models often masks fragile internal safety representations. Recent work suggests that refusal behavior is mediated by low-dimensional directions in activation space. This raises questions about how such representations are structured, localized, and accessed by optimization. We study adversarial suffix attacks as a probe of representational alignment. We introduce Activation-Guided GCG, which replaces output-based objectives with losses that directly target a model's internal refusal direction. Across several objective variants, we find that suppressing refusal globally across all layers and positions is more effective than targeting a single layer–position pair. This suggests that safety representations are distributed across the forward pass rather than causally localized to a single site. We further introduce Soft-GCG, a continuous relaxation of discrete suffix optimization using Gumbel-Softmax. Soft-GCG achieves a 33$\times$ speedup over standard GCG while improving attack success rates. Evaluating across model scales, we find that smaller models remain vulnerable while larger models resist both activation- and suffix-based attacks at our compute-constrained settings, consistent with larger and better safety trained models being harder to jailbreak. Together, our results clarify how safety mechanisms are encoded and can be broken in contemporary models. These insights provide concrete guidance for designing more robust and representation-aware alignment strategies.
\end{abstract}

\begin{links}
\link{Code}{github.com/Ege-Cakar/ImprovingGCG}
\end{links}

\section{Introduction}

The widespread deployment of Large Language Models (LLMs) has necessitated rigorous safety alignment techniques to prevent the generation of harmful or unethical content. At least in public chat interfaces, it seems like these alignment techniques have advanced significantly. However, recent research has demonstrated that these safety guardrails are surprisingly brittle. Two distinct lines of inquiry have emerged to analyze this fragility: automated adversarial attacks on the input space, and the analysis of refusal representations within the model's residual streams.

The Greedy Coordinate Gradient (GCG) method, proposed by \cite{zou2023universal}, demonstrates that optimizing adversarial suffixes can reliably bypass alignment. However, GCG operates on a behavioral proxy: it optimizes inputs to maximize the likelihood of specific target output tokens (e.g., forcing the model to output ``Sure, here is...''). While effective, this objective targets the \textit{symptom} of the jailbreak (the output logits) rather than the \textit{mechanism} of the refusal. Additionally, the GCG method is limited by the given initialization conditions and long waits until convergence. Its discrete optimization methods are significantly slower compared to continuous methods, and we hypothesize that a large portion of the optimization that is done by GCG can be done first via continuous optimization, and that discrete optimization is only necessary to overcome the ``projection gap'' (i.e., the loss in performance that comes from jumping to discrete tokens at the end after optimizing continuously and achieving a minimum in between real tokens that doesn't generalize to hard tokens).

Conversely, \cite{arditi2024refusal} investigates the model's internal representations of refusal. They demonstrated that refusal behavior in open-source chat models is mediated by a single, one-dimensional subspace in the model's activation space and that simply ablating this direction eliminates refusal behavior. This approach, though, is limited by the need for possible attacks to have white-box access to a model's activations. Thus, in its purest form, it cannot be used on closed models, most widely used by the general public.

These results raise a deeper question on how safety mechanisms are encoded internally and how robust they are to optimization. If refusal behavior is mediated by a small number of high-leverage features, then alignment may succeed behaviorally while remaining brittle at the representational level. In this paper, we use adversarial suffix optimization as a diagnostic tool for studying safety representations. Our findings suggest that current safety training produces refusal representations that are geometrically simple and accessible to gradient-based optimization. We make two contributions:
\begin{itemize}
    \item We introduce Activation-Guided GCG, which replaces output-based jailbreak objectives with losses that directly minimize projections onto internal refusal directions. By varying the spatial and depth scope of these objectives, we empirically probe whether refusal is locally or globally encoded.

    \item We introduce Soft-GCG, a continuous relaxation of discrete suffix optimization that dramatically reduces computational cost. The effectiveness of this relaxation provides evidence that the problem of the projection gap can be overcome with annealing of the ``continuity'' of the prompt during convergence.
\end{itemize}

\section{Related Work}
\label{gen_inst}
\subsection{Adversarial Attacks on Aligned LLMs}
The vulnerability of aligned LLMs to adversarial inputs has been extensively documented. Early ``jailbreaks'' relied on manual prompt engineering (e.g., ``DAN'' or role-play scenarios)\citep{shen2023do}. However, \cite{zou2023universal} introduced a formal, automated gradient-based approach known as Greedy Coordinate Gradient (GCG). GCG overcomes the discrete nature of text optimization by using token-level gradients to identify promising candidate replacements for an adversarial suffix. 

As part of their formulation, they define the adversarial loss $\mathcal{L}$ for a prompt $x_{1:n}$ as the negative log probability of a target sequence $x^*_{n+1:n+H}$:
\begin{equation}
    \mathcal{L}(x_{1:n}) = - \log p(x^*_{n+1:n+H} | x_{1:n}).
\end{equation}

For each token $x_i$ in the adversarial suffix, they compute the gradient of the loss with respect to the one-hot encoding $e_{x_i}$:
\begin{equation}
    \nabla_{e_{x_i}} \mathcal{L}(x_{1:n}) \in \mathbb{R}^{|V|}.
\end{equation}

They then select the top-$k$ candidates with the largest negative gradient values to evaluate in a forward pass. They demonstrated that this method achieves high Attack Success Rates (ASR), reaching 100\% on Vicuna-7B and 88\% on Llama-2-7B-Chat for harmful behaviors with enough iterations. Furthermore, they showed that these attacks are transferable, meaning a suffix optimized on one model can often break completely different black-box models. While highly effective, GCG is computationally extremely expensive. Additionally, it relies on the assumption that forcing the first few tokens of output determines the subsequent generation trajectory. In this paper, we aim to create a more effective suffix to bypass refusals as well as use continuous optimization to lower the computation overhead. 

\subsection{Refusal Directions}
Parallel to adversarial attacks, researchers have sought to understand how models represent refusal internally using techniques often associated with representation engineering. \cite{arditi2024refusal} utilized a difference-in-means methodology to isolate a single direction $\hat{r}$ (the normalized difference vector) in the residual stream that differentiates between harmful and harmless instructions. Concretely, they define:
\begin{equation}
    r^{(l)} = \mu^{(l)}_{\text{harmful}} - \mu^{(l)}_{\text{harmless}}
\end{equation}
where $\mu^{(l)}$ represents the mean activation at layer $l$.

They validated this finding through a technique they term \textbf{Directional Ablation}. Directional ablation erases the refusal direction from the residual stream activation $x$ by projecting it out:
\begin{equation}
    x' \leftarrow x - (x \cdot \hat{r})\hat{r}.
\end{equation}

By performing this intervention at every layer (or equivalently, orthogonalizing model weights), they effectively disabled refusal. Crucially, \cite{arditi2024refusal} compared their weight orthogonalization method directly against GCG on the HarmBench benchmark. They found their method achieved comparable or superior Attack Success Rates (ASR) while requiring significantly less computational cost as it does not require per-prompt optimization. Conversely, they demonstrated that adding this direction to harmless prompts ($x' \leftarrow x + \alpha \hat{r}$) induced refusal behavior even on benign queries like ``List the benefits of yoga'' \citep{arditi2024refusal}. Their work suggests that safety fine-tuning does not eliminate harmful knowledge but rather learns a specific ``refusal feature'' that overrides generation when activated. Our results confirm this conjecture and additionally add flexibility to their activation-based approach by developing a method to generate adversarial suffixes that optimize for bypassing refusal directions. 

\subsection{Continuous Prompt Optimization}

\cite{wen2023hardpromptseasygradientbased} proposed ``Hard Prompts Made Easy'' (PEZ), a gradient-based framework designed to bridge the optimization gap between continuous embedding spaces and discrete token sequences. Their work addresses the largest limitation in soft prompt tuning: while continuous soft prompts are easier to optimize via gradient descent, they often fail to map onto coherent or effective discrete tokens when simply projected to the nearest vocabulary neighbors after training -- a phenomenon the authors describe as the ``projection gap.'' 

To overcome this, PEZ introduces a methodology where the discretization step is integrated directly into the forward pass of the optimization loop. The algorithm maintains a set of learnable continuous parameters (soft prompts) but projects them onto their nearest discrete neighbors in the embedding matrix before computing the loss. The gradients derived from this discrete forward pass are then used to update the underlying continuous parameters. This effectively allows the optimizer to explore the smooth, differentiable continuous space while being constantly tethered to the valid discrete manifold, preventing the optimization trajectory from drifting into regions of the latent space that have no semantic validity. Empirically, Wen et al. demonstrated that this approach could robustly discover effective hard prompts for both text-to-image generation (inverting CLIP encoders for Stable Diffusion) and text-only  tasks. However, in our testing, their approach underperformed for jailbreak purposes. 

\subsection{Efficient Jailbreak Methods}

Several recent approaches have sought to reduce the computational cost of automated jailbreaks. PAIR \citep{chao2023jailbreaking} uses an attacker LLM to iteratively refine adversarial prompts based on target model responses, requiring only black-box access. TAP \citep{mehrotra2023tree} extends this with tree-of-thought reasoning and pruning for more efficient exploration. AutoDAN \citep{liu2024autodan} employs a hierarchical genetic algorithm to evolve semantically meaningful jailbreak prompts, achieving interpretability that gradient-based methods lack. These approaches trade off gradient information for query efficiency and black-box applicability. Our Soft-GCG occupies a different point in this design space: it retains white-box gradient access but replaces discrete search with continuous optimization, achieving speedups complementary to these query-based methods.

\section{Methodology}

\subsection{Activation-Guided GCG}

This approach combines the gradient-based optimization framework of GCG with mechanistic insights from refusal direction analysis. We first extract refusal directions, then use these directions to guide adversarial suffix optimization through novel activation-based objectives in the hope that we can improve performance over baseline GCG.

The activation-based approach offers two potential advantages. First, by targeting the causal mechanism underlying refusal (the internal direction identified through contrastive analysis), we may discover more fundamental adversarial perturbations than those found by optimizing surface-level output probabilities. Second, since the refusal direction is computed across many diverse examples, the activation objective may generalize better to unseen harmful prompts compared to token-level objectives that risk overfitting to specific target strings.

\subsubsection{Refusal Direction Extraction}

We use the precomputed refusal direction provided by \cite{arditi2024refusal}, extracted using the difference-in-means methodology (Equation 3). The optimal layer $\ell^* = 14$ and position $p = -1$ were selected by ranking candidate directions by how much ablating them reduces refusal scores on harmful prompts using the refusal-direction training splits. The selected direction $\mathbf{d}_{\text{refusal}}$ serves as the optimization target for our attack.

\subsubsection{Activation-Based Adversarial Optimization}

Standard GCG \citep{zou2023universal} optimizes an adversarial suffix by using one-hot representation of the tokens to compute token gradients (Equation 2). The gradients are aggregated across multiple prompts, and the top-$K$ tokens with largest negative gradients are used to generate candidate suffixes via greedy coordinate descent. We adopt this same optimization framework, replacing the optimization objective.

\subsubsection{Activation-Based Objective Functions}

The core contribution of this work is replacing the output-based loss (Equation 1) with objectives that directly target internal refusal representations. Each activation-based objective can be interpreted as a hypothesis about the structure of refusal representations. The Single objective evaluates whether refusal is causally localized to a specific layer–position pair. The Layer objective probes spatial locality across tokens at a fixed depth. The Token objective probes depth locality at a fixed position. The All objective tests whether refusal is globally distributed across the forward pass. These objectives represent different design choices in the locality-globality spectrum: from targeting a single activation, to suppressing refusal across the entire forward pass, and have significantly different optimization landscapes. Comparing optimization efficiency across these objectives allows us to assess how safety representations are structured within the model.

Given the refusal direction $\mathbf{d}_{\text{refusal}}$ at layer $\ell^*$ and position $p$ (typically the last instruction token), we explore five activation-based objectives that vary in scope and optimization target:
\begin{itemize}
    \item \textbf{Single.} Minimize projections at the most influential layer and position
\begin{equation}
    \mathcal{L}_{\text{act}}^{\text{zero}}(\mathbf{x}) = \left( \frac{\mathbf{h}^{(\ell^*)}_p(\mathbf{x}) \cdot \mathbf{d}_{\text{refusal}}}{\|\mathbf{d}_{\text{refusal}}\|} \right)^2
\end{equation}
where $\mathbf{h}^{(\ell^*)}_p(\mathbf{x})$ is the hidden state at layer $\ell^*$ and position $p$ for input $\mathbf{x} = \mathbf{x}_{\text{inst}} \oplus \mathbf{x}_{\text{adv}}$.

    \item \textbf{Negative Projection.} Push the projection to be negative (actively opposing refusal):
\begin{equation}
    \mathcal{L}_{\text{act}}^{\text{neg}}(\mathbf{x}) = \frac{\mathbf{h}^{(\ell^*)}_p(\mathbf{x}) \cdot \mathbf{d}_{\text{refusal}}}{\|\mathbf{d}_{\text{refusal}}\|}.
\end{equation}

    \item \textbf{Layer.} Minimize projections across all token positions at the most influential layer:
\begin{equation}
    \mathcal{L}_{\text{act}}^{\text{layer}}(\mathbf{x}) = \frac{1}{|\mathcal{P}|} \sum_{p \in \mathcal{P}} \left( \frac{\mathbf{h}^{(\ell^*)}_p(\mathbf{x}) \cdot \mathbf{d}_{\text{refusal}}}{\|\mathbf{d}_{\text{refusal}}\|} \right)^2
\end{equation}
where $\mathcal{P}$ represents all token positions in the sequence.

    \item \textbf{Token.} Minimize projections at a single token position across all layers:
\begin{equation}
    \mathcal{L}_{\text{act}}^{\text{token}}(\mathbf{x}) = \frac{1}{L} \sum_{\ell=1}^{L} \left( \frac{\mathbf{h}^{(\ell)}_p(\mathbf{x}) \cdot \mathbf{d}_{\text{refusal}}}{\|\mathbf{d}_{\text{refusal}}\|} \right)^2
\end{equation}
where $L$ is the total number of layers and $\mathbf{d}_{\text{refusal}}$ is the refusal direction at layer $\ell^*$, applied identically to each layer.

    \item  \textbf{All.} Minimize projections across all layers and all positions:
\begin{equation}
    \mathcal{L}_{\text{act}}^{\text{global}}(\mathbf{x}) = \frac{1}{L \cdot |\mathcal{P}|} \sum_{\ell=1}^{L} \sum_{p \in \mathcal{P}} \left( \frac{\mathbf{h}^{(\ell)}_p(\mathbf{x}) \cdot \mathbf{d}_{\text{refusal}}}{\|\mathbf{d}_{\text{refusal}}\|} \right)^2.
\end{equation}
\end{itemize}

 %\subsubsection*{Gradient Computation and Optimization}

%The gradient computation follows the same one-hot relaxation framework as standard GCG, but backpropagation terminates at the hidden state $\mathbf{h}^{(\ell^*)}_p$ rather than the output logits. This directly targets the internal refusal mechanism rather than output token probabilities. The greedy coordinate descent procedure remains unchanged, with $\mathcal{L}_{\text{act}}$ replacing the standard GCG loss in all evaluations.

\subsection{Soft-GCG}
Soft-GCG is our method of relaxing GCG into a continuous optimization problem, developed with the aim of solving the continuous problem for speedup while diverging as little as possible from GCG in terms of performance. Instead of searching for a sequence of discrete tokens (like standard GCG which swaps tokens), Soft-GCG maintains a set of continuous variables (logits) for each position in the suffix. These logits represent a probability distribution over the entire vocabulary.
\subsubsection{Problem Formulation}
Let $x$ denote the user prompt (e.g., "Write a tutorial on how to make a bomb") and $y$ denote the target response (e.g., "Sure, here is a tutorial..."). We seek an adversarial suffix $s$ of length $L$, consisting of tokens from a vocabulary $V$, such that the probability of the target response is maximized:
\begin{equation}
    s^* = \arg\max_{s \in V^L} \log P(y \mid x, s)
\end{equation}
Since the domain $V^L$ is discrete, gradients cannot be backpropagated through the token selection indices. Soft-GCG circumvents this by optimizing a continuous parameterization of the suffix.
\subsubsection{Continuous Relaxation via Gumbel-Softmax}
Instead of selecting a single discrete token $s_i$ at each position $i \in \{1, \dots, L\}$, we maintain a set of unbounded learnable logits $\phi_i \in \mathbb{R}^{|V|}$. These logits represent the unnormalized log-probabilities of selecting each token in the vocabulary.
To approximate the discrete selection process while maintaining differentiability, we utilize the Gumbel-Softmax distribution \cite{jang2017categoricalreparameterizationgumbelsoftmax}. For each position $i$, a ``soft'' token vector $\tilde{s}_i$ is computed as:
\begin{equation}
    \tilde{s}_i = \text{Softmax}\left(\frac{\phi_i + g_i}{\tau}\right)
\end{equation}
where:
\begin{itemize}
    \item $g_i \sim \text{Gumbel}(0, 1)$ are i.i.d. samples drawn from the Gumbel distribution, included to avoid convergence to local minima.
    \item $\tau > 0$ is a temperature hyperparameter controlling the spikiness of the distribution, starting smoother and allowing more cross-token interaction then pushing the vector into a one hot as it decreases.
\end{itemize}
The resulting vector $\tilde{s}_i \in \Delta^{|V|-1}$ lies on the probability simplex. The input embedding for the $i$-th suffix position, $e(s_i)$, is then approximated by the probability-weighted sum of the vocabulary embeddings matrix $E \in \mathbb{R}^{|V| \times d}$:
\begin{equation}
    e(\tilde{s}_i) = \tilde{s}_i^\top E = \sum_{v \in V} [\tilde{s}_i]_v \cdot E_v
\end{equation}
where at the end of the optimization, this should be just ``picking out'' the real token as $\tilde{s_i}$ tends towards a one-hot. This operation is fully differentiable, allowing gradients from the loss function to flow back to the logits $\phi_i$.
\subsubsection{Optimization Objectives}
\paragraph{Cross-entropy Loss} The primary objective is to minimize the negative log-likelihood of the target sequence $y$, conditioned on the prompt $x$ and the soft suffix $\tilde{s}$:
\begin{equation}
    \mathcal{L}_{CE}(\phi) = -\sum_{t=1}^{|y|} \log P(y_t \mid x, \tilde{s}, y_{<t}; \theta)
\end{equation}
where $\theta$ represents the fixed parameters of the language model. This directly aligns with the adversarial goal of maximizing target probability.

\paragraph{Carlini-Wagner Loss} In addition to the standard cross-entropy objective (Equation 13), we employ a Carlini-Wagner (CW) style loss adapted for sequence generation\citep{carlini2017evaluatingrobustnessneural}. Rather than directly maximizing the log-probability of target tokens, the CW loss optimizes for a margin between the logit of the correct target token and the highest logit among all other tokens. This encourages the model to not only rank the target token first, but to do so with a confident separation, yielding more robust suffixes upon discretization.

Formally, for a batch of target sequences with logits $z \in \mathbb{R}^{B \times T \times |V|}$, target token indices $y_t$, and an attention mask $m \in \{0,1\}^{B \times T}$ indicating valid (non-padding) positions, the per-token CW loss is:
\begin{equation}
    \ell_t^{\text{CW}} = \max\!\Big(0,\; \max_{v \neq y_t} z_{t,v} - z_{t,y_t} + \kappa\Big)
\end{equation}
where $\kappa > 0$ is a confidence margin (set to $\kappa = 5.0$ in our experiments). This formulation drives the target token logit to exceed the next-best competitor by at least $\kappa$.

To emphasize successful generation of the first target token, critical for steering the model's autoregressive trajectory, we apply a positional weighting $w_t$ with $w_0 = 5.0$ and $w_t = 1.0$ for $t > 0$. The final per-example loss, averaged over valid tokens, is:
\begin{equation}
    \mathcal{L}_{\text{CW}}(\phi) = \frac{1}{\sum_t m_t} \sum_{t=1}^{T} w_t \cdot m_t \cdot \ell_t^{\text{CW}}.
\end{equation}

We were able to establish results using both Carlini-Wagner and cross-entropy losses. 

\subsubsection{Temperature Annealing Schedules}
The temperature $\tau$ plays a critical role in the optimization trajectory. We investigate two annealing schedules:
\begin{enumerate}
    \item \textbf{Linear Schedule.} A simple decay that steadily reduces entropy over $T$ steps:
    \begin{equation}
        \tau_t = \tau_{\text{start}} - \left(\frac{t}{T}\right)(\tau_{\text{start}} - \tau_{\text{end}})
    \end{equation}
    Typically annealing from $\tau=2.0$ to $\tau=0.1$. This provides a consistent pressure towards discretization.

    \item \textbf{3-Phase ``Slushy'' Schedule.} A piecewise schedule designed to balance exploration and refinement:
    \begin{itemize}
        \item \textbf{Exploration (0\% - 50\%)}: High temperature decay ($2.5 \to 1.0$). Allows broad traversal of the embedding space.
        \item \textbf{Refinement (50\% - 75\%)}: Moderate temperature ($1.0 \to 0.5$). The ``slushy'' phase where structure forms but the distribution remains flexible enough to shift.
        \item \textbf{Solidification (75\% - 100\%)}: Rapid cooling ($0.5 \to 0.01$). Forces the soft vectors to snap to the nearest discrete tokens.
    \end{itemize}
\end{enumerate}

\subsubsection{Discretization}
After the optimization concludes, the continuous logits $\phi$ must be projected back to the discrete vocabulary to form the final usable suffix. We select the token with the highest logit value at each position:
\begin{equation}
    s_i^{\text{final}} = \arg\max_{v \in V} [\phi_i]_v
\end{equation}
This discrete suffix is then evaluated against the model to verify its effectiveness. Ideally the discretization gap is very low at this step, thanks to the latter portions of the optimization being done with an ever-decreasing temperature. 
From this point on, the converged substring can directly be utilized for attacks or it can be used as the initialization of GCG.

\section{Experimental Results}

\subsection{Activation-Guided GCG}

\subsubsection{Setup}

  All experiments use a single open-weight chat model: \texttt{Llama-2-7b-chat-hf} \citep{touvron2023llama2openfoundation}. For optimization and evaluation we rely on the datasets packaged with
  \texttt{refusal\_direction} \citep{arditi2024refusal} and \texttt{llm\_attacks} \citep{zou2023universal}:

  \begin{itemize}
    \item \textbf{Harmful prompts (train):} Both GCG and Activation-Guided GCG train on goals drawn from AdvBench \citep{zou2023universal}. For Activation-Guided GCG, the target strings are unused since the objective operates on activations rather than output logits.
    \item \textbf{Harmful prompts (test):} the harmful test split from the refusal-direction dataset, used only for evaluation.
  \end{itemize}

  We use the standard parameters provided in the \texttt{llm\_attacks} repository for all tests. Every test is trained with 200 discrete optimization steps.

  \subsubsection{Evaluation and Attack Success Metrics}

  After training a universal suffix for each method, we evaluate it on held-out
  prompts from the refusal-direction dataset.
  
  For a given method and evaluation split, we generate completions by appending
  the learned suffix to the instruction (for suffix-based methods), or by applying
  only the internal activation intervention (for ablation baselines). We then
  score each completion with three complementary methods, each inducing a notion
  of ``harmfulness''.

    The following are our evaluation definitions for harmfulness:
     \begin{itemize}
          \item \textbf{Substring ASR}:
          Labels a completion as harmful if it does \emph{not} contain any of a fixed list of refusal-style prefixes (e.g., ``I'm sorry'', ``I cannot'').
          \item \textbf{LlamaGuard2 ASR}:
          Labels a completion as harmful if it is labeled unsafe by the LlamaGuard safety classifier (\cite{metallamaguard2}).
          \item \textbf{HarmBench ASR}:
          Labels a completion as harmful if it is labeled unsafe by the HarmBench evaluation framework \citep{mazeika2024harmbenchstandardizedevaluationframework}.
      \end{itemize}

      \subsubsection{Results}
      \textbf{Activation-based objectives show improved attack success despite limited optimization steps. (Table~\ref{tab:activation_methods_asr})} As expected, the baseline (no suffix) results have the lowest ASR scores, and the ablation results are the highest. All activation-based objectives outperform or match standard GCG in substring ASR, with the ``All'' objective achieving 0.91 — approaching the ablation ceiling (0.98) — followed by Single (0.84), Layer and Negative (0.81), and Token (0.76). The strong performance of broader-scope objectives, particularly All, may reflect either that refusal representations are distributed across layers and positions, or that wider objectives provide richer gradient signal for optimization. In either case, activation-based objectives demonstrate substantially better sample efficiency than standard GCG at 200 steps.

\begin{table}[H]
    \centering
    \renewcommand{\arraystretch}{1}
    \begin{tabular}{@{}l c c c@{}} 
        \toprule
         & \multicolumn{3}{c}{\textbf{ASR (Harmful)}} \\ 
        \cmidrule(l){2-4}
        \textbf{Method} & Substring & LlamaGuard 4  & HarmBench \\
        \midrule
        Baseline           & 0.39 & 0.00 & 0.00 \\
        Ablation           & 0.98 & 0.54 & 0.62 \\
        \midrule
        Single             & 0.84 & 0.06 & \textbf{0.04} \\
        Negative           & 0.81 & 0.03 & 0.03 \\
        All                & \textbf{0.91} & \textbf{0.09} & 0.01 \\
        Layer              & 0.81 & 0.03 & 0.01 \\
        Token              & 0.76 & 0.00 & 0.00 \\
        GCG      & 0.76 & 0.00 & 0.00 \\
        \bottomrule
    \end{tabular}
    \caption{
        Harmful prompt ASR for suffix-based attacks and refusal-direction ablation on the refusal-direction \texttt{harmful\_test} split. We report results using the suffix with the lowest loss during training.
    }
    \label{tab:activation_methods_asr}
\end{table}

\subsection{Soft-GCG}

\begin{figure*}[ht]
    \centering
    \includegraphics[width=0.8\linewidth]{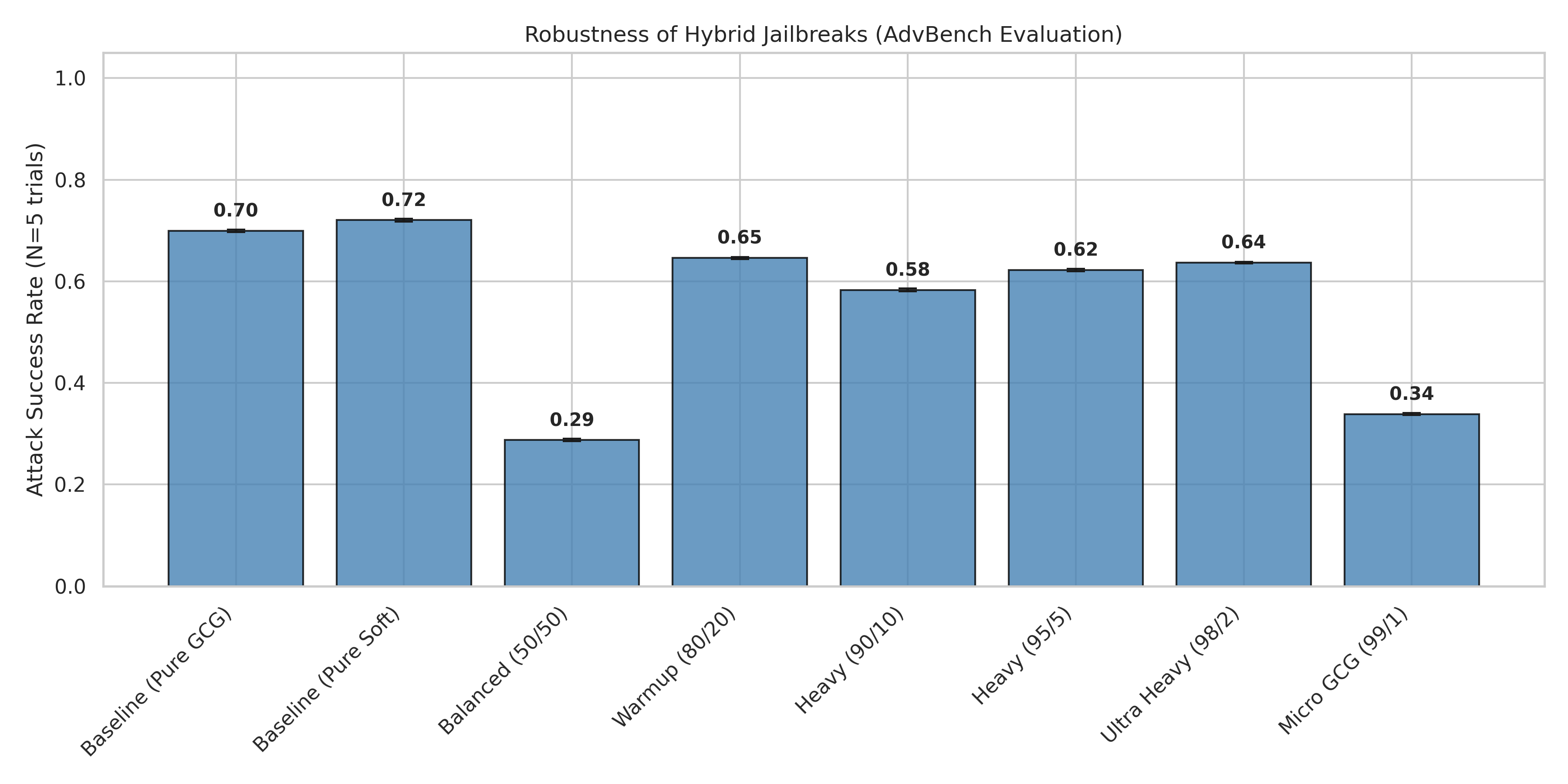}
    \caption{Performance of different configurations of SGCG$\rightarrow$GCG. We see that it is more useful to run more SGCG iterations than GCG (number of iterations are constant).}
    \label{fig:soft_gcg}
\end{figure*}

\textbf{SGCG by itself achieves performance parity with GCG in our experiments, showing no need for any pure GCG steps at the end. (Figure~\ref{fig:soft_gcg})} Given the computational cost of GCG, we generate one suffix per configuration for our long-main runs. The resulting suffixes are then evaluated against the entire AdvBench corpus multiple times to accurately assess performance. We observe similar performance parity between SGCG and GCG in smaller-scale runs. All results are evaluated on prompts from AdvBench using Llama-2-7b-chat-hf and substring ASR, with the same substrings as in the original GCG paper. For sweep experiments, we use a split of 25 training prompts and 10 test prompts, all drawn from AdvBench.
\begin{figure}
    \centering
    \includegraphics[width=0.75\linewidth]{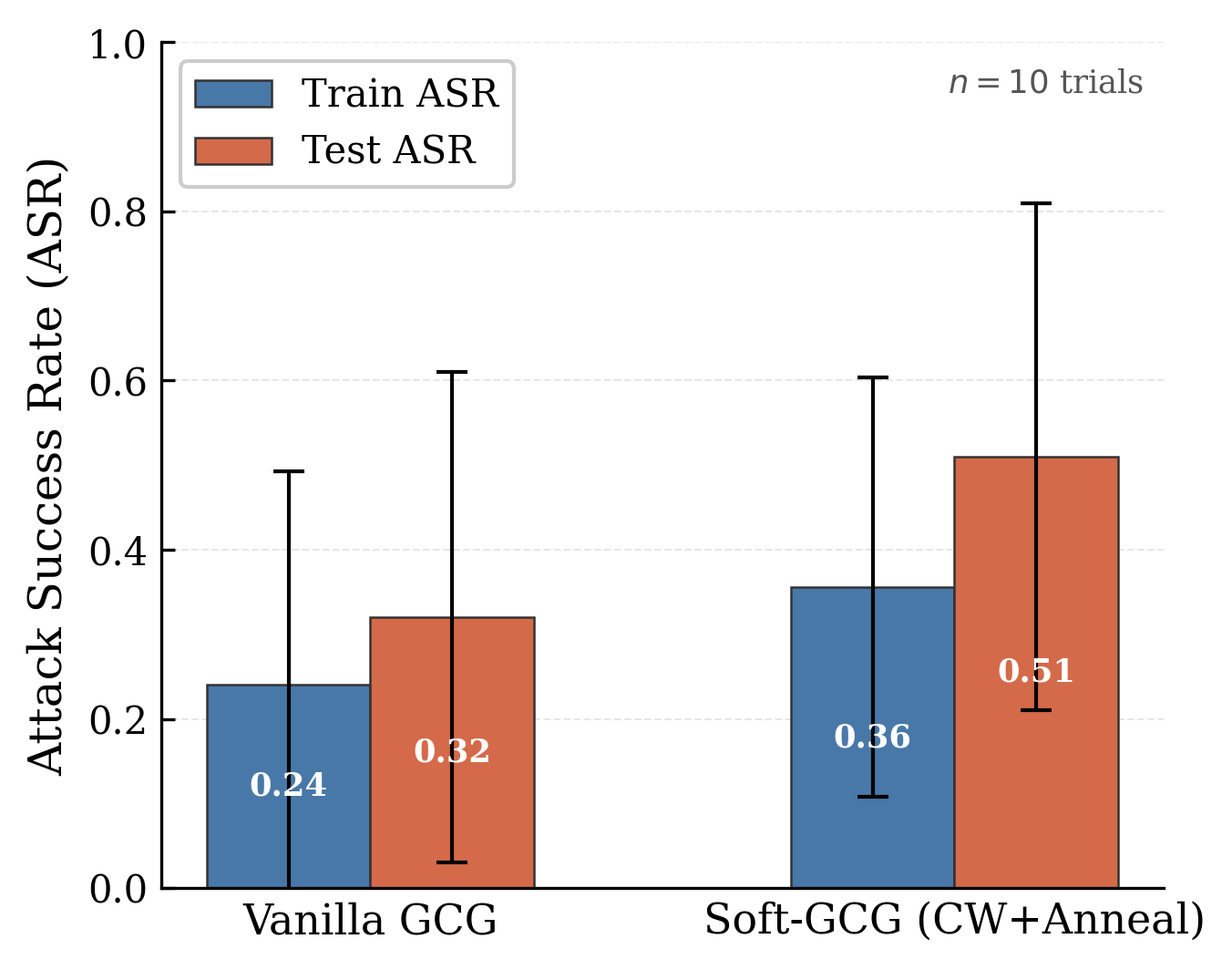}
    \caption{Comparison of pure Soft-GCG and GCG. With $\sim 33\times$ speed, S-GCG beats vanilla GCG in performance as well. ASR was calculated through the substring method here. }
    \label{fig:sgcg-vgcg}
\end{figure}
\subsubsection{Optimization Setup and Configuration}

The suffix length is fixed at $L=20$ tokens. In sweep experiments, we varied combinations of Soft and GCG steps, including Pure GCG (500 steps), Pure Soft (500 steps), and hybrid configurations (e.g., 250/250, 400/100). For the Gemma experiments, we ran 500 steps of Pure Soft optimization informed by sweep results. We use a learning rate of 0.1 for the Adam optimizer applied to suffix logits, a batch size of 10 for Gemma experiments, and 128 GCG candidates in the sweep. For the extended seed-comparison experiments (Figure~\ref{fig:sgcg-vgcg}), we increased the iteration count to 2000.

SGCG's primary advantage is computational efficiency: it completes in approximately 2.5 minutes compared to 81 minutes for vanilla GCG on the same hardware---a \textbf{33$\times$ speedup}---while also achieving higher ASR. In the shorter 500-step runs the speedup reached $\sim$43$\times$, though we found 2000 steps yielded more consistent results. Since the main bottleneck is model loading rather than per-iteration cost, SGCG scales favorably to longer optimization runs.

We then evaluate SGCG on the Gemma 3 model family, selected for two reasons: (a) Ollama, one of the most popular local serving applications, identifies it as the strongest model family that can run on a single GPU \citep{Ollama}, and (b) it originates from a major AI lab (Google) with wide adoption \citep{Google_Gemma3_2025}, making it a realistic target for local deployment.

\begin{table}[h]
    \centering
    \renewcommand{\arraystretch}{1}
    \begin{tabular}{@{}l c@{}}
        \toprule
        \textbf{Model} & \textbf{Substring ASR$^*$} \\
        \midrule
        Gemma3-270m & 1.000 $\pm$ 0.000 \\
        Gemma3-1b   & 0.577 $\pm$ 0.031 \\
        Gemma3-4b   & 0.336 $\pm$ 0.104 \\
        Gemma3-12b  & 0.000 $\pm$ 0.000 \\
        \bottomrule
    \end{tabular}
    \caption{
        ASR of the Gemma3 family. The 27B model was not evaluated given the negligible ASR observed on the 12B variant. 
        ASR is marked with an asterisk because the 270m model became incoherent after optimization.
    }
    \label{tab:sgcg_gemma}
\end{table}

In Table~\ref{tab:sgcg_gemma}, the observed drop in attack success with increasing model size suggests a scaling-dependent change in how refusal is represented. Larger models may encode safety constraints in higher-rank or more distributed subspaces, reducing the effectiveness of both localized activation suppression and smooth input-space optimization. This points toward a potential representational phase transition in alignment mechanisms as models scale, with implications for robustness and interpretability.
\section{Discussion and Future Work}

We develop two complementary contributions exposing critical vulnerabilities in aligned language models: (1) a novel white-box attack targeting internal refusal representations, demonstrating that safety mechanisms encoded as linear directions can be systematically suppressed via adversarial optimization, and (2) a 33x speed improvement to GCG that reduces computational barriers dramatically while increasing performance. By achieving high attack success rates on modern models (up to 34\% ASR on Gemma 3 4B), we provide concrete evidence that current RLHF-based alignment is insufficient against representation-level attacks. The speed improvement transforms attacks from resource-intensive operations into ones executable within minutes on consumer hardware, making this a practical exploit accessible to broader actors. We demonstrate success on frontier models, and we intend for this to be an address to foundation model developers, in that Soft-GCG allows for longer suffixes to be trained for much longer for jailbreaking larger models. We chose not to run those out of safety concerns and lack of compute, but that is not necessarily the case for adversarial parties.

Our theory relies on the security mindset: publicly disclosing attack methodologies -- including efficiency improvements -- accelerates defense development faster than exploitation. We assume (1) white-box access requirements limit immediate harm despite increased speed, (2) currently exploitable models aren't yet smart enough to be destructive, (3) AI labs will prioritize robustness when vulnerabilities are demonstrated to be effective and rapidly exploitable, and (4) insights from current models remain relevant for future architectures. The 33x speedup demonstrates that computational protections are shallow, compelling faster defensive action. The ultimate impact is next-generation AI systems developed with representation-level robustness: safety mechanisms distributed beyond exploitable linear directions, adversarial training including efficient activation-based attacks, and red-teaming testing for mechanistic vulnerabilities discoverable at scale. The primary risk is that findings improve attacks rather than defenses before adequate countermeasures are deployed.

\subsection{Limitations}

Even though we argue that Gemma 3 is the most suitable model for evaluation, it does come with its problems. The most significant one is the huge vocabulary size of 262 thousand for all Gemma 3 models \citep{Google_Gemma3_2025}, as opposed to $\sim$128 thousand for Llama 3 \citep{HuggingFace_Llama3_2024}. Due to the way we parametrize SGCG, the vocabulary size directly increases the complexity of the problem, making it a harder optimization. This, however, is entangled with the safety training of specific models, so we were unable to isolate the effect of vocabulary size directly on performance.  

There are also limitations in how we evaluate our methods. Particularly, we use substring matching to calculate ASR for both Activation-Guided GCG and Soft-GCG. Substring matching can lead to both false negatives and false positives, since it isn't adaptable to the response itself and solely looks for common strings like ``I cannot'', etc to determine refusal. Future work could involve using more robust metrics of evaluation, such as using LlamaGuard as a judge for Soft-GCG.

\section{Safety Statement}
This work makes jailbreaking attacks significantly more efficient, and we are transparent about the dual-use implications. A speedup reduces the computational cost of circumventing alignment, which matters beyond controlled research settings, particularly as capable open-weight models become more widely accessible.

The attacks currently require white-box access to weights and gradients, but we do not think this access barrier should be treated as a lasting safeguard.

We believe this work should be published because the vulnerabilities we exploit are not introduced by our methods; they are intrinsic to how current alignment techniques work. Efficient attacks are necessary for rigorous red-teaming, and safety researchers need these tools to audit weaknesses before adversaries independently develop them. More fundamentally, our results suggest that alignment representations learned through current training are fragile under gradient-based optimization, and we hope this motivates investment in more robust alignment strategies.

Mitigating these vulnerabilities likely requires moving beyond surface-level alignment fine-tuning. Promising directions include adversarial training directly over the representation space targeted by our attacks, circuit-level interventions that harden the specific mechanisms underlying refusal behavior, and simply training against these types of attacks. Ultimately, we view the efficiency gains demonstrated here as raising the bar for what counts as a credible robustness evaluation -- defenses should be validated against fast, gradient-based attacks of this class before being considered reliable.

\bibliography{aaai2026}

@article{zou2023universal,
  title={Universal and Transferable Adversarial Attacks on Aligned Language Models},
  author={Zou, Andy and Wang, Zifan and Carlini, Nicholas and Nasr, Milad and Kolter, J. Zico and Fredrikson, Matt},
  journal={arXiv preprint arXiv:2307.15043},
  year={2023}
}

@inproceedings{arditi2024refusal,
  title={Refusal in Language Models Is Mediated by a Single Direction},
  author={Arditi, Andy and Obeso, Oscar and Syed, Aaquib and Paleka, Daniel and Panickssery, Nina and Gurnee, Wes and Nanda, Neel},
  booktitle={Advances in Neural Information Processing Systems},
  volume={38},
  year={2024}
}

@article{shen2023do,
  title={``Do Anything Now'': Characterizing and Evaluating In-the-Wild Jailbreak Prompts on Large Language Models},
  author={Shen, Xinyue and Chen, Zeyuan and Backes, Michael and Shen, Yun and Zhang, Yang},
  journal={arXiv preprint arXiv:2308.03825},
  year={2023},
  url={https://arxiv.org/abs/2308.03825}
}

@misc{wen2023hardpromptseasygradientbased,
      title={Hard Prompts Made Easy: Gradient-Based Discrete Optimization for Prompt Tuning and Discovery}, 
      author={Yuxin Wen and Neel Jain and John Kirchenbauer and Micah Goldblum and Jonas Geiping and Tom Goldstein},
      year={2023},
      eprint={2302.03668},
      archivePrefix={arXiv},
      primaryClass={cs.LG},
      url={https://arxiv.org/abs/2302.03668}, 
}

@misc{jang2017categoricalreparameterizationgumbelsoftmax,
      title={Categorical Reparameterization with Gumbel-Softmax}, 
      author={Eric Jang and Shixiang Gu and Ben Poole},
      year={2017},
      eprint={1611.01144},
      archivePrefix={arXiv},
      primaryClass={stat.ML},
      url={https://arxiv.org/abs/1611.01144}, 
}

@misc{ollama,
  title        = {Ollama -- Gemma 3},
  author       = {{Ollama}},
  year         = {2024},
  howpublished = {\url{https://ollama.com/library/gemma3}},
  note         = {Accessed: 2025-09-30}
}

@misc{HuggingFace_Llama3_2024,
  author       = {Philipp Schmid and Omar Sanseviero},
  title        = {Welcome Llama 3 – Meta’s new open LLM},
  howpublished = {\url{https://huggingface.co/blog/llama3}},
  year         = {2024},
}

@misc{Google_Gemma3_2025,
  author       = {Google Developers Blog},
  title        = {Gemma explained: What's new in Gemma 3},
  howpublished = {\url{https://developers.googleblog.com/en/gemma-explained-whats-new-in-gemma-3/}},
  year         = {2025},
}

@misc{chao2023jailbreaking,
    title={Jailbreaking Black Box Large Language Models in Twenty Queries},
    author={Patrick Chao and Alexander Robey and Edgar Dobriban and Hamed Hassani and George J. Pappas and Eric Wong},
    year={2023},
    eprint={2310.08419},
    archivePrefix={arXiv},
    primaryClass={cs.LG}
}

@misc{mehrotra2023tree,
    title={Tree of Attacks: Jailbreaking Black-Box LLMs Automatically},
    author={Anay Mehrotra and Manolis Zampetakis and Paul Kassianik and Blaine Nelson and Hyrum Anderson and Yaron Singer and Amin Karbasi},
    year={2023},
    eprint={2312.02119},
    archivePrefix={arXiv},
    primaryClass={cs.LG}
}

@inproceedings{liu2024autodan,
    title={AutoDAN: Generating Stealthy Jailbreak Prompts on Aligned Large Language Models},
    author={Xiaogeng Liu and Nan Xu and Muhao Chen and Chaowei Xiao},
    booktitle={The Twelfth International Conference on Learning Representations},
    year={2024},
    url={https://openreview.net/forum?id=7Jwpw4qKkb}
}

@misc{touvron2023llama2openfoundation,
      title={Llama 2: Open Foundation and Fine-Tuned Chat Models}, 
      author={Hugo Touvron and Louis Martin and Kevin Stone and Peter Albert and Amjad Almahairi and others},
      year={2023},
      eprint={2307.09288},
      archivePrefix={arXiv},
      primaryClass={cs.CL},
      url={https://arxiv.org/abs/2307.09288}, 
}

@misc{metallamaguard2,
  author={Hakan Inan and Kartikeya Upasani and Jianfeng Chi and others},
  title={Llama Guard: LLM-based Input-Output Safeguard for Human-AI Conversations},
  year={2023},
  howpublished = {\url{https://github.com/meta-llama/PurpleLlama/blob/main/Llama-Guard2/MODEL_CARD.md}},
}

@misc{mazeika2024harmbenchstandardizedevaluationframework,
      title={HarmBench: A Standardized Evaluation Framework for Automated Red Teaming and Robust Refusal}, 
      author={Mantas Mazeika and Long Phan and Xuwang Yin and Andy Zou and Zifan Wang and Norman Mu and Elham Sakhaee and Nathaniel Li and Steven Basart and Bo Li and David Forsyth and Dan Hendrycks},
      year={2024},
      eprint={2402.04249},
      archivePrefix={arXiv},
      primaryClass={cs.LG},
      url={https://arxiv.org/abs/2402.04249}, 
}

@INPROCEEDINGS{carlini2017evaluatingrobustnessneural,
  author={Carlini, Nicholas and Wagner, David},
  booktitle={2017 IEEE Symposium on Security and Privacy (SP)}, 
  title={Towards Evaluating the Robustness of Neural Networks}, 
  year={2017},
  volume={},
  number={},
  pages={39-57},
  doi={10.1109/SP.2017.49}}
\end{document}